\documentclass[letterpaper, 10 pt, conference]{ieeeconf}
\IEEEoverridecommandlockouts
\usepackage{cite}
\usepackage{amsmath,amssymb,amsfonts}
\usepackage{algorithmic}
\usepackage{graphicx}
\usepackage{textcomp}
\usepackage{xcolor}
\def\BibTeX{{\rm B\kern-.05em{\sc i\kern-.025em b}\kern-.08em
    T\kern-.1667em\lower.7ex\hbox{E}\kern-.125emX}}
\begin{document}

\title{Is 3D Convolution with 5D Tensors Really Necessary for Video Analysis?\\
}

\author{Habib Hajimolahoseini$^{1}$, Walid Ahmed$^{1}$, Austin Wen$^{1}$, and Yang Liu$^{1}$
\thanks{$^{1}$ Ascend Team, Toronto Research Centre, Huawei Technologies, Toronto, Canada
        {\tt\small habib.hajimolahoseini@huawei.com}}%
}


\maketitle

\begin{abstract}
In this paper, we present a comprehensive study and propose several novel techniques for implementing 3D convolutional blocks using 2D and/or 1D convolutions with only 4D and/or 3D tensors. Our motivation is that 3D convolutions with 5D tensors are computationally very expensive and they may not be supported by some of the edge devices used in real-time applications such as robots. The existing approaches mitigate this by splitting the 3D kernels into spatial and temporal domains, but they still use 3D convolutions with 5D tensors in their implementations. We resolve this issue by introducing some appropriate 4D/3D tensor reshaping as well as new combination techniques for spatial and temporal splits. The proposed implementation methods show significant improvement both in terms of efficiency and accuracy. The experimental results confirm that the proposed spatio-temporal processing
structure outperforms the original model in terms of speed and accuracy using only 4D tensors with fewer parameters.
\end{abstract}


\section{Introduction}
During the past few years, 3D convolutional neural networks have become dominant in the area of video analysis, especially for action recognition \cite{carreira2017quo, feichtenhofer2019slowfast, feichtenhofer2020x3d, wang2018non, lin2019tsm, shao2020temporal, li2020tea, zhang2020v4d, ryoo2019assemblenet, yang2020temporal, varol2017long, hajimolahoseini2012extended}. 
However, 3D convolution is computationally very expensive, which may cause problems in real-time applications. 
That is mostly because the common strategy in video processing models is to expand a well-known 2D image architecture into a 3D spatio-temporal model \cite{hajimolahoseini2008improvement}. 
For example, C3D \cite{tran2015learning} and I3D \cite{carreira2017quo} are the 3D versions of VGG-16 \cite{simonyan2014very} and Inception-V1 \cite{szegedy2015going}, respectively. 
Although this type of architectures could reach the state-of-the-art in accuracy, it has been proved that this 2D to 3D expansion approach is not optimal in terms of computational costs \cite{feichtenhofer2020x3d}.
For example, the 3D version of ResNet used in video processing uses around 27 times more mathematical operations than its 2D version used in image recognition \cite{he2016deep, wang2018non}. 

In order to increase the efficiency of deep learning models, different strategies have been proposed in the literature \cite{ahmed2023speeding, ataiefard2024skipvit, javadi2023gqkva, hajimolahoseini2024single, hajimolahoseini2024methods, hajimolahoseini2021compressing, hajimolahoseini2023methods, hajimolahoseini2023training, hajimolahoseini2023swiftlearn}. 
In video processing applications, early attempts were mostly focused on using 2D convolutions with some tricks in order to include the motion analysis as well. 
Two-stream networks are examples of this type of methods in which one stream is used for spatial processing while motion analysis is performed in the second stream using 2D convolutions \cite{simonyan2014two, wang2016temporal}.
However, the computational costs added by the optical flow calculation in the motion stream prevent these methods from being efficient in real-time applications. 
On the other hand, there are some other 2D based networks e.g. TSM \cite{lin2019tsm} and TIN \cite{shao2020temporal} that perform the motion analysis by shifting some of the features along the temporal dimension in order to provide the ability of information exchange between the adjacent frames. 

Another straightforward approach is to split the network architecture by performing the 2D convolutions in some layers of the network for spatial processing while applying the 3D convolutions for spatio-temporal analysis in the rest of the model. 
In \cite{xie2018rethinking}, it is shown that efficiency and accuracy could be improved by applying 2D convolutions to the early layers of the network to extract high-level semantic information from frames.
The temporal representation learning is then performed at the top of the network by applying 3D convolutional layers to these high-level features. 
ECO architecture is one of the most famous networks in this category \cite{zolfaghari2018eco}. 

A similar strategy is to use different pathways for spatial and temporal analysis as proposed in the SlowFast architecture \cite{feichtenhofer2019slowfast}. 
In that network, the spatial processing is performed in the Slow pathway using more features at a lower frame rate, while the temporal analysis is done in the Fast pathway using less features at a higher frame rate. 
They then extended their work to a single path architecture called X3D, in which a tiny 2D network is progressively expanded in different dimensions to a 3D architecture instead of just adding an extra temporal dimension to a 2D image based network \cite{feichtenhofer2020x3d}. 

On the other hand, instead of splitting the network into 2D and 3D layers or different pathways, another alternative solution is to factorize each 3D layer into spatial and temporal domains throughout the network. 
In this approach, every 3D convolutional block is decomposed into a spatial 2D convolution followed by a temporal 1D convolution. 
S3D \cite{xie2018rethinking}, P3D \cite{qiu2017learning} and R(2+1)D \cite{tran2018closer} are some of the well-known architectures in this category. 
As presented in R(2+1)D architecture, one benefit of this approach is that the network capacity could be doubled by adding an extra non-linear function in between the 2D and 1D convolutions while preserving the number of parameters \cite{tran2018closer}. 
This could lead to a higher accuracy and efficiency comparing to the original 3D convolutional networks.

Factorization of the 3D blocks can go even deeper by decomposing the kernels into a consecutive sequence of one-dimensional filters across all directions in order to improve the efficiency even more \cite{jin2014flattened}.  
However this implies a strong assumption that the convolutional kernels are of rank-1 so that the matrix decomposition is reversible by cross production of all 1D components \cite{kim2018rank}.  

On the other hand, some networks take advantage of channel-wise separable convolutions or group convolutions.
In CSN architecture for example, it is shown that separating channel and spatio-temporal interactions could improve both efficiency and accuracy at the same time as it acts as a regularization technique \cite{tran2019video}.  
In that method, all 3D convolutional blocks are split into a point-wise ($1\times1\times1$) convolution for inter-channel processing and a depth-wise ($3\times3\times3$) convolution for spatio-temporal analysis inside each channel. 

In video analysis, the input to the network is a 5D tensor with the following shape: 
\begin{center}
  Tensor Shape = [$B$, $T$, $X$, $Y$, $C$]
\end{center}
in which $B$, $T$, $X$, $Y$ and $C$ represent batch size, number of frames, width, height and number of channels, respectively.
In 3D convolution, although the kernels are assumed to be 3 dimensional, in reality the following 4D filter is applied to the input tensors:
\begin{center}
  Filter Shape = [$t$, $w$, $h$, $c$]
\end{center}
where $t$, $w$, $h$ and $c$ are the dimensions of filters in time, horizontal, vertical and channel domains. 
Note that in regular 3D convolutions, the kernel's channel dimension $c$ is equal to the number of channels in the input tensor $C$ ($c=C$). 
Therefore, all the features are collapsed into a single channel and this process is repeated until the desired number of output channels is achieved. 
On the other hand, in group convolutions, channels are divided into different groups and all features within a group are collapsed into a single channel. 
Finally in depth-wise convolution, no interaction between the channels is performed and so it is assumed that $c=1$, which means that in each input channel, a 3D filter of shape $t\times w\times h$ is applied independently \cite{dumoulin2016guide}. 
For the sake of simplicity in presentations, we omit the channel dimension of the filters and assume 3D kernels of shape $t\times w\times h$ throughout the rest of this paper. 

Although in theory, the above-mentioned techniques simulate the 3D convolution using a combination of 2D and/or 1D convolutions, in their implementation codes they still use 3D convolutions with 5D tensors, but with specific parameters for spatial and temporal analysis. 
In other words, instead of applying a 3D convolution with a $t\times h\times w$ kernel to the 5D input tensor of shape $B\times T\times X\times Y\times C$, 3D simulation approaches apply two different 3D convolutions to the 5D tensors, one with a $1\times w\times h$ kernel and another one with a $t\times1\times1$ kernel for spatial and temporal analysis, respectively. 
While this approach of modeling the 3D convolutions may work when training on powerful multipurpose systems, it may cause complications and limitations for some of the edge devices.   
Furthermore, in almost all of the existing methods, the spatial and temporal convolutions are applied in a sequential form in which, temporal analysis is applied to the output of the spatial analysis. 
However, other combination options e.g. parallel spatial and temporal processing and different ways of combining them e.g. by summation, concatenation, etc. are not fully studied. 

In this paper, we study and propose some new alternative operations for modelling the  3D convolutions using either 2D convolutions with 4D tensors, 1D convolution with 3D tensors, or a combination of them. 
This is done by splitting the 4d filters used in the 3D convoluitons into spatial and temporal domains with appropriate reshaping and combinations of the tensors. 
For each method, different ways of combining spatial and temporal analysis i.e. sequential, parallel, summation, concatenation and etc.,are also explored in order to find the most efficient and accurate combination method. 
The proposed simulation techniques show significant improvement both in terms of efficiency and accuracy.

\section{Proposed Method}
\begin{figure}[h]
    \centering
    \includegraphics[width=0.5\textwidth]{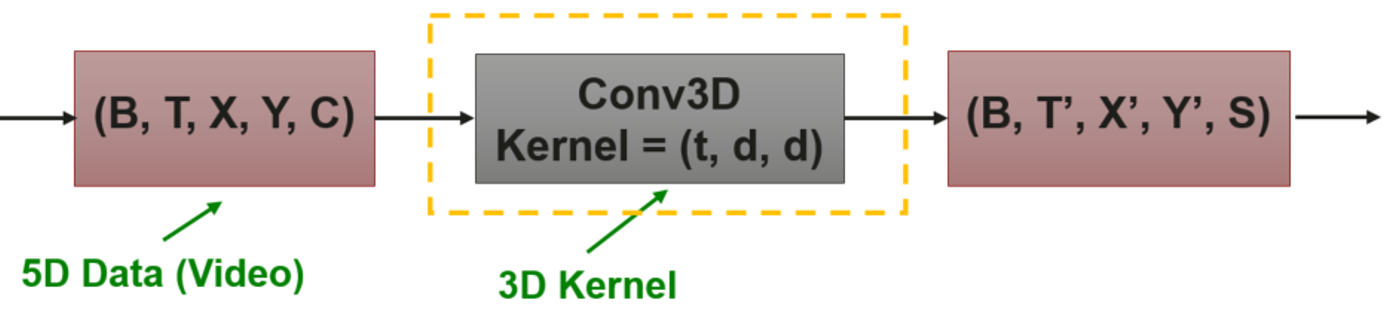}
    \caption{A regular 3D convolutional layer applied to 5D tensors.}
    \label{org}
\end{figure}

A regular 3D convolutional layer and its 5D input/output shapes are shown in Fig.\ref{org}. 
As shown in this figure, it applies a 3D convolution operation to both special ($X$, $Y$) and temporal dimensions ($T$) at the same time to extract the spatio-temporal information. 
Note that in almost all of the video analysis models, spatial dimensions $w$ and $h$ of the kernels are the same. 
Therefore, we assume: $w=h=d$ for the sake of simplicity in representations. 
The output would have the shape of $B\times T' \times X' \times Y' \times S$, where $T'$, $X'$ and $Y'$ are the new temporal and spatial dimensions and $S$ is the number of output channels.

\begin{figure}[h]
    \centering
    \includegraphics[width=0.4\textwidth]{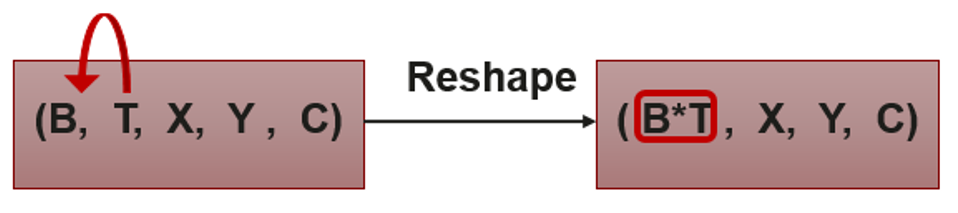}
    \caption{Reshaping the input 5D tensor into 4D.}
    \label{reshape4d}
\end{figure}
\begin{figure*}[h]
    \centering
    \includegraphics[width=1\textwidth]{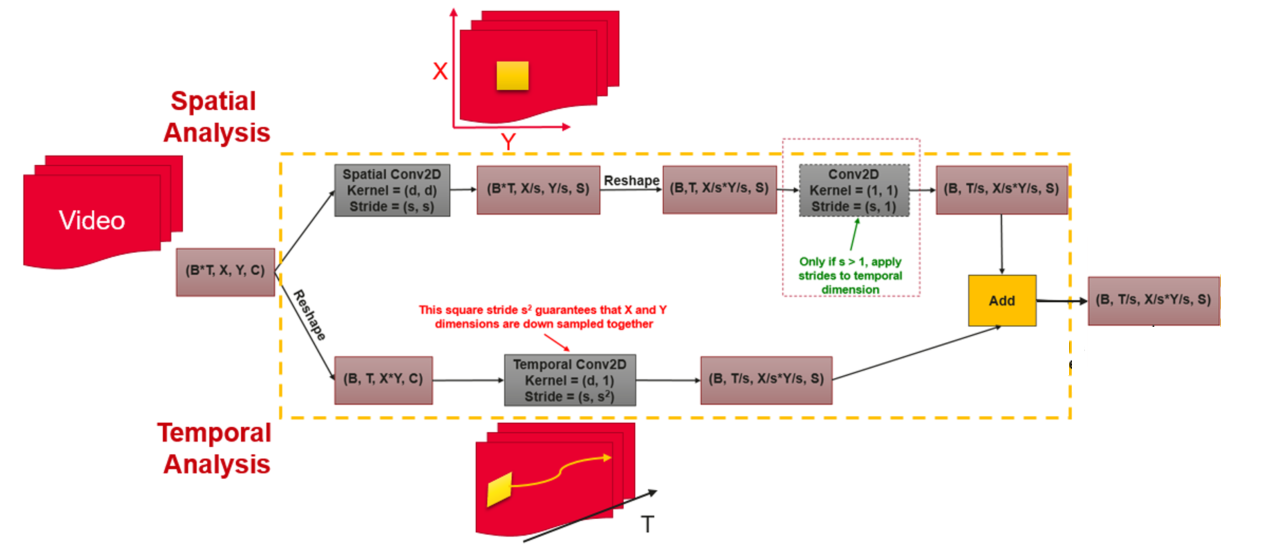}
    \caption{The proposed architecture which replaces the regular 3D convolutional layers.}
    \label{proposed}
\end{figure*}

In order to avoid using 5D tensors, we first reshape the input data into 4D format by multiplying its first 2 dimensions as shown in Fig.\ref{reshape4d}.
Then, in order to avoid applying 3D operations, we replace the 3D convolutional layer in Fig.\ref{reshape4d} with the proposed structure shown in \ref{proposed}. 

As shown in Fig.\ref{proposed}, we apply the spatial and temporal processing independent from each other in two parallel branches: spatial branch which analyses the data in its $X$ and $Y$ dimensions, and temporal branch which analyses the input in its temporal domain $T$. 

\subsection{Spatial Analysis}
Each frame of the video can be considered as a static image with spatial 2D data. 
In the spatial analysis branch, the first 2D convolution applies the spatial analysis to each frame by multiplying the $X$ and $Y$ dimensions with a $d\times d$ kernel. 
This will generate a 4D tensor of size: $[B\times T, \frac{X}{s}, \frac{Y}{s}, S]$ in which the lowercase “s” is an integer representing the stride. 
After the spatial analysis is finished, in order to keep the tensor dimension under 4D, we now multiply the number of vertical and horizontal pixels of each frame by reshaping the tensor into: $[B, T, \frac{X}{s}\times \frac{Y}{s}, S]$. 
After that, a 1-by-1 convolution with stride of $s$ is applied to the temporal dimension (if necessary) in order to pool the temporal dimension by $s$. 
The output of the spatial branch will then have the following size: $[B, \frac{T}{s}, \frac{X}{s}\times \frac{Y}{s}, S]$. 

\subsection{Temporal Analysis}
On the other hand, in the temporal analysis branch, we only analyze the relationship between each pixel of the images in consecutive frames. 
To do this, the input tensor is first reshaped by multiplying the vertical and horizontal pixels of the frames: $[B, T, X\times Y, C]$.
Then, a 2D convolution with kernel $d\times 1$ and strides of $(s, s^2)$ is applied to the 2nd and 3rd dimensions of the 4D tensor which performs both the temporal analysis and spatial striding at the same time. 
The reason behind using $s^2$ as our spatial stride is that the X and Y dimensions are multiplied and applying an $s^2$ stride to the 3rd dimension will shrink each of X and Y dimensions by a factor of s. 
Therefore, this square stride guarantees that the output of the temporal branch will have the exact same size as that of the spatial branch. 
This will make combining the output of two spatial and temporal branches very straightforward by simply adding them together. 
The resulting tensor generated by adding of two branches outputs will also have the same size as each of the branch outputs: $[B, \frac{T}{s}, \frac{X}{s}\times \frac{Y}{s}, S]$.

\section{Experimental Results}
We choose ECO-Lite architecture (with a minor modification) as the baseline in our experiments \cite{zolfaghari2018eco}. 
This architecture is composed of two parts: a set of 2D layers called 2D-Net for spatial analysis of individual frames, and a set of 3D layers called 3D-Net for spatio-temporal analysis of the feature representations learned from the 2D-Net.
The first few layers of Inception-V3 \cite{szegedy2016inception} (originally BN-Inception \cite{ioffe2015batch} in the ECO paper) is used as the 2D-Net while the last few layers of 3D-Resnet18 \cite{tran2017convnet} is adopted for the 3D-Net. 
The modified ECO-Lite architecture is presented in Table \ref{eco-lite}.

\begin{table}
\begin{center}
\begin{tabular}{|c|c|c|}
\hline
Layer Name & Output Size & Filters \\
\hline\hline
\textbf{2D-Net}\\
\hline\hline
conv2D-1 & $121\times121$ & $[3\times3, 32]$ \\
\hline
conv2D-2 & $119\times119$ & $[3\times3, 32]$ \\
\hline
conv2D-3 & $119\times119$ & $[3\times3, 64]$ \\
\hline
pool1 & $59\times59$ & $[3\times3]$ \\
\hline
conv2D-4 & $59\times59$ & $[1\times1, 80]$ \\
\hline
conv2D-5 & $57\times57$ & $[3\times3, 192]$ \\
\hline
pool2 & $28\times28$ & [$3\times3$] \\
\hline
inception (3a) & $28\times28$ & $[- 256]$ \\
\hline
Inception (3b) & $28\times28$ & $[- 288]$ \\
\hline
Inception (3c) & $28\times28$ & $[- 96]$ \\
\hline\hline
\textbf{3D-Net}\\
\hline\hline
conv3D-1 & $N\times28\times28$ & $\begin{bmatrix}
3\times3\times3, 128 \\
3\times3\times3, 128 
\end{bmatrix}\times2$ \\
\hline
conv3D-2 & $\frac{N}{2}\times14\times14$ & $\begin{bmatrix}
3\times3\times3, 256 \\
3\times3\times3, 256 
\end{bmatrix}\times2$ \\
\hline
conv3D-3 & $\frac{N}{4}\times7\times7$ & $\begin{bmatrix}
3\times3\times3, 512 \\
3\times3\times3, 512 
\end{bmatrix}\times2$ \\
\hline
pool3 & $1\times1\times1$ & $[\frac{N}{4}\times7\times7]$ \\
\hline
fc, softmax & $1\times1\times1$ &  [$1\times1\times1$, \#classes]\\
\hline
\end{tabular}
\end{center}
\caption{The modified ECO-Lite architecture, assuming that the input is a video clip with $N$ frames of size $243\times243$ \cite{zolfaghari2018eco}}
\label{eco-lite}
\end{table}

There are multiple reasons for choosing the ECO-Lite architecture in our experiments. 
First of all, it is simple for implementation and efficient during both training and inference. 
Second of all, almost half of the architecture consists of 2D convolutional layers from Inception-V3 which we do not need to change during our experiments with different 3D simulations techniques in the 3D-Net. 
This will provide us with a huge benefit in terms of training time as we can initialize the 2D-Net using the Inception-V3 pre-trained weights on a large image dataset e.g. ImageNet \cite{deng2009imagenet}. 
This will also enable us to transfer knowledge between our different 3D simulation experiments as we only change the 3D-Net during our experiments while 2D-Net stays untouched. 

Experiments are performed on NVIDIA V100 GPUs. 
In the experiments, all 3D Convolutional layers in ECO-Lite architecture are replaced by one of the equivalent structures in literature that resemble the 3D convolution including: R(2+1)D \cite{tran2018closer}, P3D-A, -B, and -C \cite{qiu2017learning} and Rank-1 \cite{kim2018rank}. 
Two versions of the proposed structure are also implemented, in which the last block in Fig.\ref{proposed} (the yellow add block) that combines the temporal and spatial branches are adding (Proposed-add) or concatenation (Proposed-cat) operations.  
Concatenation is done along the last dimension of the tensors (channels). 

The number of training parameters, floating points operations per second (FLOPs) and inference speed in terms of frame per second (FPS) of all of the structures are also compared with that of the baseline Conv3D module in Table \ref{inference-speed}.   
As seen in this table, the proposed method with adding block (Prop-Add) has only 16.3 million parameters which is 51\% less than the baseline (Conv3D). 
the FLOPs also drops by 51\% and the inference speed improves by 12\%, which is the highest speed-up among all other implementations. 

\begin{table}
\begin{center}
\begin{tabular}{c|c|c|c}
Method & \#Params(M) & FPS & FLOPs(G)\\
\hline\hline
Conv3D  & 33.7 & 773 & 35.6\\
\hline\hline
R(2+1)D  & 33.7 & 764 & 36.9\\
\hline
Proposed-Cat  & 32.2 & 774 & 32.7\\
\hline
Proposed-Add  & 16.3 & 865 & 17.1\\
\hline\hline
P3D-A  & 22.5 & 824 & 19.5\\
\hline
P3D-B  & 23.5 & 791 & 20.9\\
\hline
P3D-C  & 22.8 & 758 & 19.6\\
\hline\hline
Rank-1  & 13.7 & 724 & 14.7\\
\end{tabular}
\end{center}
\caption{Comparison of the efficiency of different structures resembling the Conv3D module}
\label{inference-speed}
\end{table}

In order to evaluate the performance of different structures, we use two datasets including Kinetics-400 \cite{kay2017kinetics} and UCF-101 \cite{soomro2012ucf101}. 
The experimental results for the different structures on these datasets are shown in Table \ref{accuracy-ucf-kinetics}. 
In this table, the 3D layers of ECO-Lite architecture are trained from scratch while the 2D blocks are initialized using the ImageNet weights of Inception-V3. 
The performance results of pretraining on Kinetics dataset and then finetuning on UCF is also reported in Table \ref{accuracy-kinetics-ucf}. 
In this experiment, the model is first trained on Kinetics dataset for 50 epochs and then finetuned on UCF for 100 epochs. 

\begin{figure}[h]
    \centering
    \includegraphics[width=0.5\textwidth]{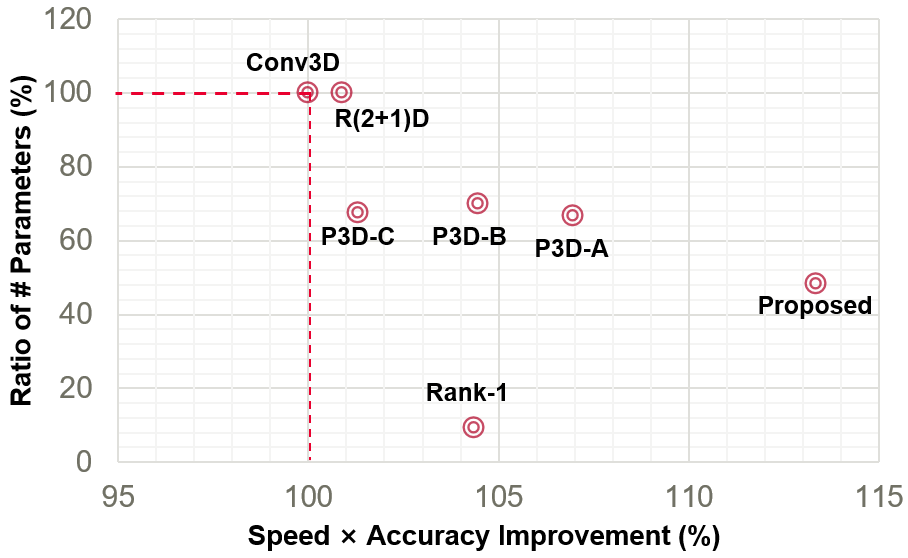}
    \caption{Number of parameters vs. speed and accuracy improvement for different methods. Note that the lower the number of parameters and the higher the speed and accuracy, the better the model.}
    \label{acc_vs_speed}
\end{figure}

As seen in these tables, the proposed technique is more efficient and even more accurate than the other methods, even the baseline 3D-CNN which uses 5D tensors. 
More specifically, the Proposed-Cat method has the highest accuracy on both Kinetics and UCF datasets in both cases, even better than the baseline. 
The possible reason could be the higher non-linearity caused by the new structures which increases the capacity of the networks. 
If the number of parameters is not an issue, the Proposed-Cat structure is the best choice among all of the techniques. 
The Proposed-Add shows the highest speed-up among all of the techniques although it does not have the highest accuracy. 
However, it still preserves the accuracy higher than the baseline. 
Therefore, it could be a good choice when the speed and efficiency is a priority. 
This fact is shown in Fig.\ref{acc_vs_speed}, in which the compression ratio is depicted as a function of the speed multiplied by accuracy improvement. 
\begin{table}
\begin{center}
\begin{tabular}{c|c|c|c}
Method & 2D-Net Pre & Kinetics & UCF \\
\hline\hline
Conv3D & ImageNet & 55.63 & 72.85 \\
\hline\hline
R(2+1)D & ImageNet & 57.91 & 69.00 \\
\hline
Proposed-Cat & ImageNet & 59.41 & 70.15 \\
\hline
Proposed-Add & ImageNet & 57.90 & 75.79 \\
\hline\hline
P3D-A & ImageNet & 56.77 & 67.88 \\
\hline
P3D-B & ImageNet & 57.83 & 69.62 \\
\hline
P3D-C & ImageNet & 59.40 & 71.56 \\
\hline\hline
Rank-1 & ImageNet & 56.14 & 64.12 \\
\end{tabular}
\end{center}
\caption{Accuracy comparison when training the 3D layers from scratch on Kinetics-400 and UCF-101, using ImageNet weights for the 2D layers}
\label{accuracy-ucf-kinetics}
\end{table}


\begin{table}
\begin{center}
\begin{tabular}{c|c|c}
Method & Pretrained & Top-1(\%) \\
\hline\hline
Conv3D  & Kinetics & 88.48 \\
\hline\hline
R(2+1)D  & Kinetics & 90.31 \\
\hline
Proposed-Cat  & Kinetics & 91.21 \\
\hline
Proposed-Add  & Kinetics & 90.62 \\
\hline\hline
P3D-A  & Kinetics & 90.44 \\
\hline
P3D-B  & Kinetics & 90.31 \\
\hline
P3D-C  & Kinetics & 91.01 \\
\hline\hline
Rank-1  & Kinetics & 86.41 \\
\end{tabular}
\end{center}
\caption{Accuracy comparison when training the model on Kinetics-400 for 50 epochs and then fine-tuning on UCF-101}
\label{accuracy-kinetics-ucf}
\end{table}

\section{Conclusion}
In this work, we studied some techniques for implementing the 3D convolutional layers using 2D and/or 1D convolutions with only 4D and/or 3D tensors.
  The existing approaches reshapes the 5D tensors at the begigning of the models and analyses the data in two different branches including spatial and temporal domains. 
  We performed this by introducing some appropriate 4D/3D tensor reshaping as well as new combination techniques for spatial and temporal splits. 
  The proposed implementation methods show significant improvement both in terms of efficiency and accuracy. 
  We have performed multiple experiments on both NVIDIA's GPUs as well as Huawei's Ascend AI accelerators.  
In summary we solved the existing problems with conventional 3D-CNN as follows:
\begin{itemize}
    \item Appropriate reshaping techniques in the parallel structure of the proposed method allows us to use 4D tensors throughout the entire system instead of 5D
\item	Also, the 4D tensors in both branches will have the same shapes after processing which makes them more efficient to combine by adding
\item	Using only 2D kernels enables us to implement the spatial and temporal processing much more efficiently with significantly lower memory consumption, which are critical in real-time applications especially for edge devices in real-world applications. 

\end{itemize}

{\small
\bibliography{citations}
\bibliographystyle{style_ref}
}

\end{document}